\documentclass[10pt,twocolumn,letterpaper]{article}

\usepackage{iccv}
\usepackage{times}
\usepackage{epsfig}
\usepackage{graphicx}
\usepackage{amsmath}
\usepackage{amssymb}
\usepackage{subcaption}
\usepackage{float}
\newsavebox{\largestimage}

 \usepackage[pagebackref=true,breaklinks=true,letterpaper=true,colorlinks,bookmarks=false]{hyperref}

\iccvfinalcopy 

\def\iccvPaperID{4200} 

\ificcvfinal\fi
\begin{document}

\title{Face Video Generation from a Single Image and Landmarks}

\author{\parbox{16cm}{\centering
    {\large Kritaphat Songsri-in$^1$, Stefanos Zafeiriou$^{1,2}$}\\
    {$^1$ Imperial College London, $^2$ University of Oulu}}\\
    {\tt\small \{kritaphat.songsri-in11, s.zafeiriou\}@imperial.ac.uk}
}


\maketitle

\begin{abstract}
In this paper we are concerned with the challenging problem of producing a full image sequence of a deformable face given only an image and generic facial motions encoded by a set of sparse landmarks. To this end we build upon recent breakthroughs in image-to-image translation such as pix2pix, CycleGAN and StarGAN which learn Deep Convolutional Neural Networks (DCNNs) that learn to map aligned pairs or images between different domains (\ie, having different labels) and propose a new architecture which is not driven any more by labels but by spatial maps, facial landmarks. 
In particular, we propose the MotionGAN which transforms an input face image into a new one according to a heatmap of target landmarks.  We show that it is possible to create very realistic face videos using a single image and a set of target landmarks. Furthermore, our method can be used to edit a facial image with arbitrary motions according to landmarks (\eg, expression, speech, \etc). This provides much more flexibility to face editing, expression transfer, facial video creation, \etc. than models based on discrete expressions, audio or action units.
\end{abstract}

\section{Introduction}
The problem of editing and manipulating faces in images and videos has countless applications spanning from post-production of movies and dubbing to generation of synthetic results for facial expression recognition and lipreading. Until recently, this problem belonged to the field of computer graphics and was solved by building person-specific models \cite{3dmm, face2face} or manual editing. Currently, due to the advent of machine learning and in particular with the introduction of Generative Adversarial Networks (GANs) \cite{gan}, the problem is now re-designed using GAN architectures. 
\begin{figure}[t]
\begin{center}
  \includegraphics[width=\linewidth]{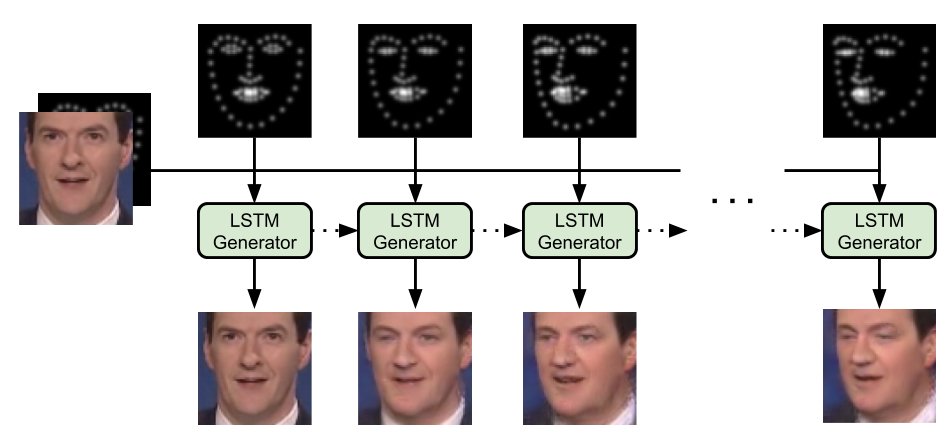}
\end{center}
  \caption{Our framework generates a video based on a single given source image and target sparse landmarks.}
\label{fig:firstfig}
\end{figure}
In computer vision terms, the problem of face editing falls in the domain of image-to-image translation under different settings. In the most straightforward case an image-to-image translation method is modelled as a conditional-GAN (cGAN) \cite{cgan} trained with aligned image pairs. In the absence of image pairs but presence of labels (\ie, domains) cycle-consistency losses have  been used to train image-to-image translation models (\eg, CycleGAN \cite{CycleGAN2017}, StarGAN \cite{stargan}, \etc.). In this context, StarGAN architecture can be used to train a model for facial expression transfer using discrete expressions. Following the same line of research as StarGAN, the recent GANimation \cite{ganimation} translates a facial image according to the activation of certain facial Action Units (AUs) \footnote{AUs is a system of taxonomy for classifying motions of the human facial muscles \cite{au}.} and their intensities. Even though AU is a quite comprehensive model for describing facial motion, detecting AUs is currently an open problem both in controlled \cite{conau}, as well as in unconstrained recording conditions \cite{auishard1, auinvid}. In particular, in unconstrained conditions for certain AUs the detection accuracy is not high enough yet \cite{auinwild}. One of the reasons is the lack of annotated data owing to the high cost of annotation which has to be performed by highly trained experts. Finally, AUs cannot describe all possible lip motion patterns produced during a speech. Hence, the GANimation model cannot be used in a straightforward manner for transferring speech.  Other recent methods such as \cite{dinobaseline, dino} generate synthetic speech videos condition to audio. Nevertheless, the results in \cite{dinobaseline} look unrealistic, since only the mouth region moves, while the method in \cite{dino} is only applied to a limited dataset. The recent X2Face method \cite{x2face} proposes to drive face synthesis using codes produced by embedding networks (\eg, face recognition, pose \etc.). Nevertheless, the method assumes that certain variations are fully disentangled, which is not always the case. Furthermore, the generated images lack high-frequency details, probably owing to the lack of the use of an adversarial training strategy.

In this paper, we are motivated by the remarkable results achieved by recent methodologies in facial landmarks localization and tracking \cite{zafeiriou2017menpo,zafeiriou20173d,guo2018dual} and propose the first, to the best of our knowledge, face image and video synthesis methodology driven by facial landmarks. Contrary to StarGAN and GANimation, our methodology does not need any human annotations, as we operate using pseudo-annotations provided by state-of-the-art facial landmarks localisation algorithms \cite{jiankangfg2018}. An overview of our method is shown in Figure \ref{fig:firstfig}. We meticulously designed an image-to-image translation methodology with adversarial training that does not suffer from error accumulation, so that it is suitable for video generation. In summary the contributions of our work are 
\begin{itemize}
    \item We designed a generator that takes a facial image, its current landmarks, and target landmarks as inputs and generates a facial image with the same identity but with target landmarks. 
    \item We proposed an exhaustively designed adversarial training architecture for our generator that is not prone to drift owing to error accumulation. That is, we incorporated adversarial losses both for an image, as well as a video generation. That way, our generator can be applied to generate images and videos. 
    \item For identity preservation and correct motion transfer we included both verification, as well as landmark localisation losses. 
\end{itemize}

\section{Related works}

{\bf Generative Adversarial Networks (GANs)} are a class of generative models. A GAN usually consists of two competing networks: a generator and a discriminator. The discriminator's goal is to distinguish between real and generated samples while the generator tries to produce examples as realistic as possible in order to fool the discriminator. The competition between the two networks influences the generator to produce more realistic and less blurry results. The original framework in \cite{gan} was developed for image generation from normally distributed random noises, but the framework has been adopted by the community to tackle various other problems such as cGAN and image-to-image translations. Due to the popularity of the framework, there are several GAN-based works \cite{began, wgan, cvae},  that extend the method and improve upon image quality as well as the stability during training.

{\bf Image-to-Image Translations}: Most of the recent methods that perform the task of image-to-image translations capitalize on the power of GANs. In the case that paired data points are presented, pix2pix \cite{pix2pix} performs image translation between two domains based on the $\ell_1$ loss and an adversarial loss. However, when only labels between two domains are available, CycleGAN instead exploits the cycle consistency losses between them. Nevertheless, these methods are not scalable especially for image translations on multiple domains since two pairs of generator and discriminator are needed for each possible domain translation. Recently, StarGAN proposed to solve this problem by using only a single generator. Compared to prior methods, StarGAN enables multiple domains translation by fusing target domain attributes with the given image by concatenating them channel-wise.

{\bf Video-to-Video Translations}: There are several works that focus on video-to-video generations. Inspired by CycleGAN, RecycleGAN \cite{Recycle-GAN} translates video contents between two specific domains. In addition to cyclic consistency loss, RecycleGAN imposes spatio-temporal constraints between creating realistic results between two seen video domains. Focusing on face-related frameworks, X2Face proposed to synthesise videos based on a learned face representation image extracted from a sequence of source identity videos. Based on driving videos or other conditions such as head poses or audio inputs, the method generates a sequence of driving vectors which in turn move the embedded face image to produce a target video.  Another interesting work on video-to-video translations that is based on sparse landmarks is DyadGAN which generates face expressions in dyadic interactions. The method proposes to produce the video of the interviewer based on the video of the interviewee. The framework consists of two stages, one to generate sketched images of the target domains from the source domain, and the other to generate face images based on the sketch.


\section{Proposed Framework}
We describe our face video synthesis network that generates a sequence of realistic face video frames $\widetilde{f}^T_1 \equiv [\widetilde{f}_1, \widetilde{f}_2, ..., \widetilde{f}_T]$ based on a given source image $s$, its landmark $l$, and a sequence of target landmarks $l^T_1 \equiv [l_1, l_2, ..., l_T]$. We interpret 2D landmarks positions as heat-maps images where each channel represents each landmark locations. At each channel, the position of each landmark is described by a 2D Gaussian distribution whose mean peak at the ground truth position (see Figure \ref{fig:firstfig}). 


\subsection{Sub Networks}
\begin{figure*}
    \centering
    \savebox{\largestimage}{\includegraphics[width=0.26\textwidth]{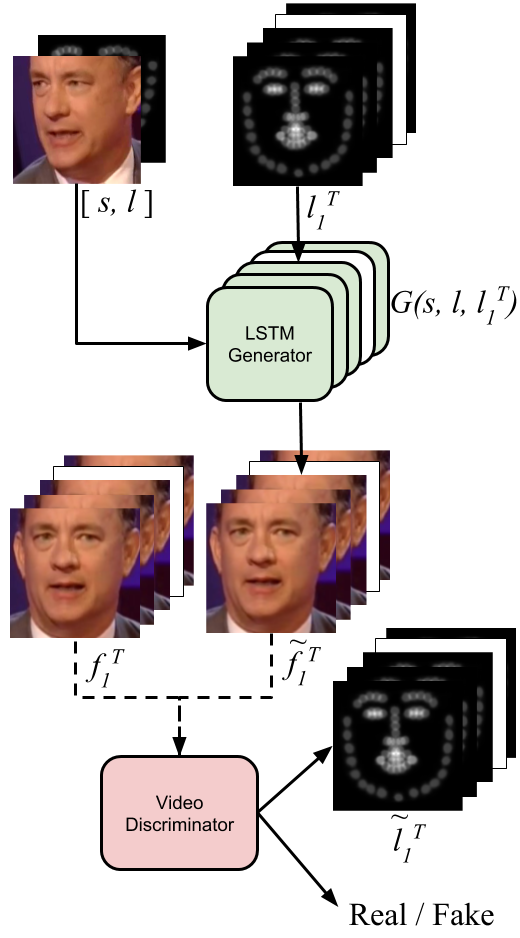}}
    \begin{subfigure}[b]{0.33\textwidth}
        \centering
        \includegraphics[width=\textwidth]{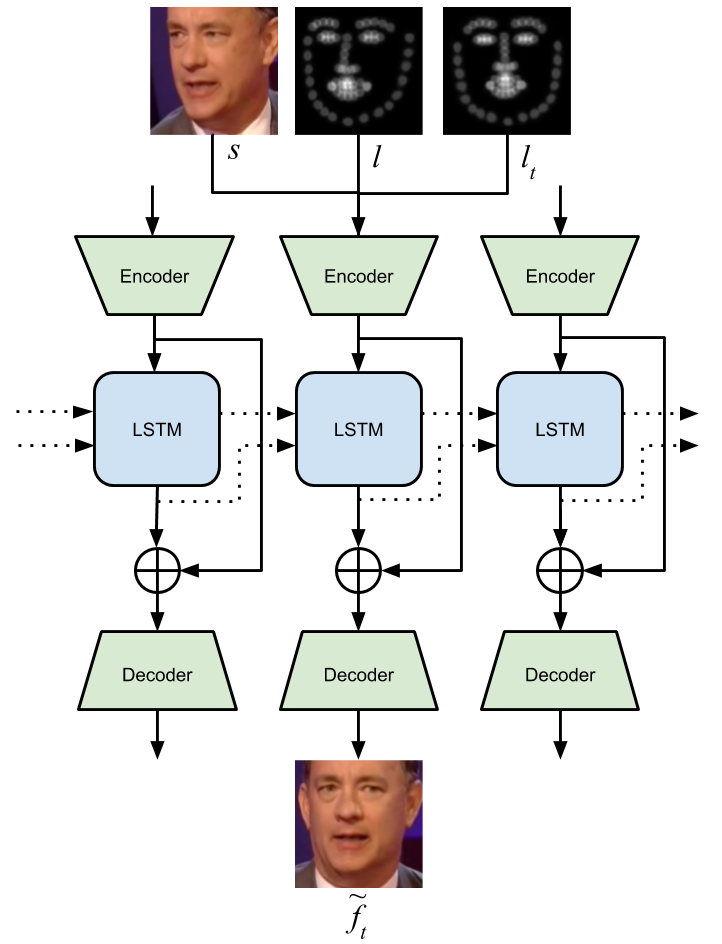}
        \caption{\label{fig:generator}Our generator consists of three parts: an encoder, a LSTM network, and a decoder.}
    \end{subfigure}
    ~ 
    \begin{subfigure}[b]{0.36\textwidth}
        \centering
        \raisebox{\dimexpr.5\ht\largestimage-.5\height}{
        \includegraphics[width=\textwidth]{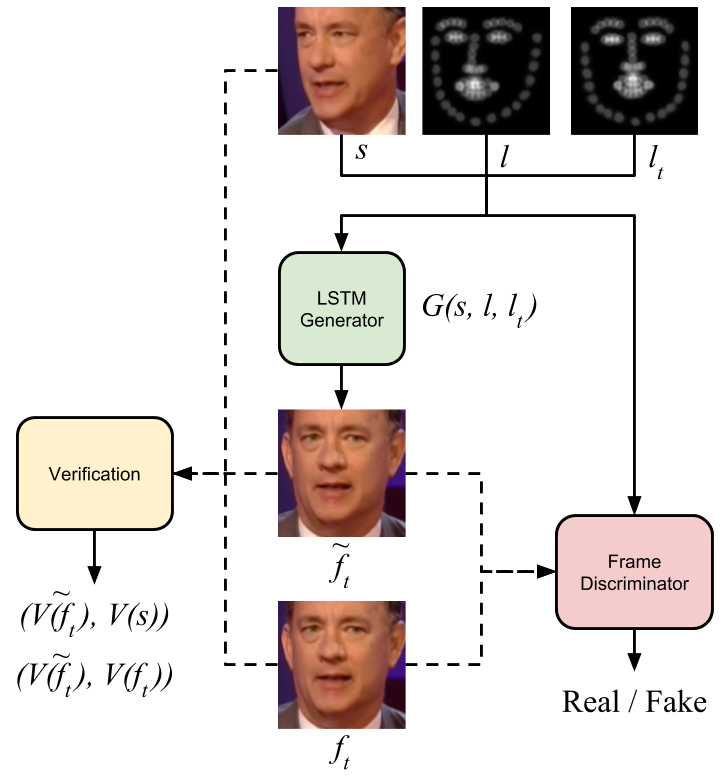}}
        \caption{\label{fig:frame}A frame generation and its associated face verification and frame discriminator networks.}
    \end{subfigure}
    ~ 
    \begin{subfigure}[b]{0.26\textwidth}
        \centering
        \usebox{\largestimage}
        \caption{A video generation utilizes a video discriminator network.}
    \end{subfigure}
    \caption{\label{fig:video}Our overall networks for a facial video generation.}
    \label{fig:overall}
\end{figure*}

Our framework is based on Generative Adversarial Networks which contains four networks: a generator $G$, an image frame discriminator $D_f$, a video discriminator $D_v$, and a verification network $V$. Every network in our model is shown in Figure \ref{fig:overall}.

{\bf Generator:} 
As seen in Figure \ref{fig:generator}, our generator is based on a Long Short-Term Memory network (LSTM) \cite{lstm}. The input of the generator $G$ is a channel-wise concatenation of source image, its landmarks, and target landmarks $[s, l, l_t]$. Our generator network architecture is based on \cite{stargan} which consists of two convolution down-sampling layers followed by six layers of residual modules, and finally ends with two transpose convolution up-sampling layers. We split the network in \cite{stargan} into two parts: the first half for an encoder and the second half for a decoder. The output of the encoder is passed through a single layer LSTM block. The output of the LSTM block is then element-wisely added to the output of the encoder and become the decoder input. The output of the decoder is a facial image $\widetilde{f}_t = G(s, l, l_t)$.
For brevity of the notation, we omit cell and hidden states internally used by the LSTM and define a video generation for $T$ consecutive frame as
\begin{equation}
    \widetilde{f}_1^T = G(s, l, l_1^T)
\end{equation}
Using a LSTM-based generator allows us to synthesize videos sequentially frame-by-frame. During training, we force the landmarks of the source image $l$ and the first target landmark $l_1$ to not come from consecutive frames. With this setting, we can also use our LSTM-based generator to produce a single target image based on a target landmark $l_1$ as shown later in our experiments section.

{\bf Frame Discriminator:} 
$D_f$ takes either an image $f_t$ or a generated frame $\widetilde{f}_t$ concatenated with a source image, its landmarks, and target landmark as $[s, l, f_t, l_t]$ or $[s, l, \widetilde{f}_t, l_t]$ as inputs. The architecture of $D_f$ is also the same as that of the generator $G$ but without the LSTM unit. We also follow the output layer of CycleGAN which used a patch discriminator to perform prediction of local areas. As a result, the output is now an image with one channel representing the prediction of the discriminator. The final prediction is then the average value of predictions of every location. Since we conditioned the input on source image $s$ and its landmark $l$, not only will the frame discriminator $D_f$ distinguishes between real and generated frame but it will also recognize whether the identity and other attributes of the generated frame $\widetilde{f}_t$ is preserved. 

{\bf Video Discriminator:}
$D_v$ takes a stack of $T$ real video frames $f^{T}_1$ or generated video frames $\widetilde{f}_1^{T}$ as inputs. It's architecture is also similar to the generator $G$'s without the LSTM unit. However, in order for the video discriminator $D_v$ to encapsulate the temporal movement of the video, we replace 2d convolution layers used in the generator $G$ with 3d convolution layers while we keep others layers as well as other activation functions the same. Besides, the network $D_v$ also has two branches at the up-sampling layers: one to predict if a given video sequence is real or fake, and the other one to output the corresponding landmarks $\widetilde{l}^{T}_1 \equiv D^l_v(f_1^{T}) \; \text{or} \; D^l_v(\widetilde{f}_1^{T})$ depending on the given video input. As a result, the generator $G$ is encouraged to produce realistic videos whose landmarks are similar to target landmarks $l_1^{T}$. 

{\bf Verification Network:}
The verification network $V$ is a pre-trained network that was trained for face recognition problem in \cite{lightcnn}. The network is kept fixed throughout the training procedure and only used to ensure that the generated frames $\widetilde{f_t}$ preserve the identity of the source image $s$. In order to utilize all the available identity information, we use $V$ to compute face features of two pairs: between generated frame and source image $(V(\widetilde{f}_t), V(s))$, and between generated frame and target frame, $(V(\widetilde{f}_t), V(f_t))$. These pairs of features are then later used to compute identity loss described in the following subsection.


\subsection{Loss functions}
\subsubsection{Image Reconstruction Loss}
Considering each individual generated frame from our generator $\widetilde{f}_t = G(s, l, l_t)$, we adopt pixel-wise $\ell_1$ norm as a reconstruction loss for the generator $G$ as:
\begin{equation}
    L^G_{\textit{img}}= \frac{1}{T} \sum\limits_{t=1}^T || G(s, l, l_t) - {f}_t ||
    \label{eq:rec loss}
\end{equation}
where $f_t$ is the corresponding ground truth frame whose landmarks is $l_t$.
\subsubsection{Adversarial Loss}
It is well-known that that using only reconstruction loss in \ref{eq:rec loss} produces blurry images. To further improve the quality of the generated frames and produce realistic looking results, we adopted two discriminators in our framework: a frame generator $D_f$ and a video discriminator $D_v$. It was shown in \cite{dualdis, multidis} that multiple discriminators can lead to faster and more stable training. In our case, we tailored each discriminator to solve different aspects occurred in our problem. 

{\bf Frame Adversarial Loss:}
The frame discriminator $D_f$ is used to ensure that each generated frame $\widetilde{f}_t$ looks realistic and preserves the identity of the source image $s$. We deploy adversarial loss to the frame discriminator $D_f$ for each frame of the given generated video $\widetilde{f}_1^{T}$ and the target video $f_1^{T}$ which is defined as:
\begin{align}
    \begin{split}
        L^{D_f}_{\textit{adv}}= 
    & \frac{1}{T} \sum\limits_{t=1}^T \;
     \mathbb{E}_{f_t}[\log(D_f(s, l, f_t, l_t))] + \\ & \mathbb{E}_{l_t}[\log(1 -D_f(s, l, G(s, l, l_t), l_t))]
    \label{eq:frame adv}
    \end{split}
\end{align}

{\bf Video Adversarial Loss:}
The video discriminator $D_v$ encourages the generator $G$ to produce realistic videos which follow target landmarks $l_1^{T}$.  We deploy adversarial loss to the video discriminator $D_v$ as:
\begin{align}
\begin{split}
    L^{D_v}_{\textit{adv}}= \; &\mathbb{E}_{f_1^{T}}[\log(D_v(f_1^{T}))]  + \\ &\mathbb{E}_{l_1^{T}}[\log(1 - D_v(G(s, l, l_1^T)))]
    \label{eq:video adv}
\end{split}
\end{align}

{\bf Pairwise Feature Matching Loss:}
Vanilla GAN's generator loss tends to have a problem with vanishing gradient which leads to unstable training or mode collapse \cite{wgan, began}. Here, we adapt feature matching loss from \cite{cvae} which was shown to provide more stable training and produce realistic results. Unlike \cite{gan, began, wgan}, the feature matching loss exploits the availability of real target images or videos. The pairwise feature matching loss between $\widetilde{f}_1^{T}$ and ${f}_1^{T}$ for the generator $G$ against both frame discriminator $D_f$ and video discriminator $D_v$ is defined using $\ell_2$ norm as :
\begin{align}
\begin{split}
    L^G_{\textit{adv}}= & \frac{1}{T} \sum\limits_{t=1}^T || I_{D_f}(G(s, l, l_t)) - I_{D_f}({f}_t) ||^2_2 + \\ &
    || I_{D_v}(G(s, l, l_1^{T})) - I_{D_v}({f}_1^{T}) ||^2_2
    \label{eq:gen loss df}
\end{split}
\end{align}
where $I_{D_f}$ and $I_{D_v}$ are the intermediate layers of the frame discriminator $D_f$ and the video discriminator $D_v$ respectively.

\subsubsection{Landmarks Reconstruction Loss}
Our video discriminator $D_v$ is specifically designed to also output predicted landmarks $\widetilde{l}_1^{T} \equiv D^l_v(f_1^{T}) \; \text{or} \; D^l_v(\widetilde{f}_1^{T})$ depending on the given video input. Not only does our video discriminator $D_v$ optimize the video adversarial loss $L^{D_v}_{\textit{adv}}$ defined in $\ref{eq:video adv}$ but also the landmark reconstruction loss for target video frames $f_1^T$ using $\ell_2$ norm defined as:
\begin{equation}
    L^{D_v}_{\textit{lms}}= ||D^l_v(f_1^{T}) - {l}_1^{T}) ||^2_2
    \label{eq:gen loss df}
\end{equation}
On the other hand, the generator $G$ is encouraged to produce video frames $\widetilde{f_1^T}$ that follow target landmarks $l_1^T$ by optimizing its landmarks reconstruction loss which is:
\begin{equation}
    L^{G}_{\textit{lms}}= ||D^l_v(G(s, l, l_1^{T})) - {l}_1^{T}) ||^2_2
    \label{eq:gen loss df}
\end{equation}

\section{Experiments}
In this section, we demonstrate the capability of our framework by presenting both qualitative and quantitative results on three tasks: facial video synthesis from a facial image and target landmarks, facial video synthesis based on audio inputs, and changing facial image emotion.

\subsection{Implementation Details}
We followed an alternating training loop among the generator $G$, the frame discriminator $D_f$, and the video discriminator $D_v$. The generator's parameters are optimized against $\lambda_1L^G_{img} + \lambda_2L^G_{adv} + \lambda_3L^G_{lms} + \lambda_4L^G_{id}$. The frame discriminator's parameters are optimized against $L^{D_f}_{adv}$, and the video discriminator's parameters are optimized against $\lambda_5L^{D_v}_{adv} + \lambda_6L^{D_v}_{lms}$. Based on the validation set, the values of these hyper-parameters are: $\lambda_1 = 1, \lambda_2 = 0.01, \lambda_3 = 10, \lambda_4 = 0.1, \lambda_5 = 1$, and $\lambda_6 = 100$. Each of these network is consecutively trained with adaptive moment estimation optimizer (ADAM) \cite{adam} with parameters: $\alpha=0.0001, \beta_1=0.9$, and $\beta_2=0.999$. Lastly, due to memory limitation, we truncated our video sequences to be $T=4$ \ie the number of stacked frames fed to the generator and the video discriminator.

\subsection{Video Generation from Target Landmarks}
\subsubsection{Dataset}
For this task, we train and evaluate our model with a facial video database, 300VW \cite{DBLP:conf/iccvw/ShenZCKTP15, Chrysos_2015_ICCV_Workshops} which consists of 114 videos and around 218k frames. The dataset is annotated with 68 landmarks and includes videos in arbitrary conditions. Since the dataset contains face images at various scales, we crop each frame based on the ground truth landmarks and resize them to $128\times 128$. During training, we augment the dataset by mirroring the video sequence. At testing, videos are generated based on the first frame of each video.
\subsubsection{Baselines}
Since there is no prior work that performs image-to-image translation on a given face image into corresponding target face conditioned on target landmarks, we consider two possibles candidate frameworks that could be adopted as our baselines: CycleGAN and StarGAN.

{\bf CycleGAN}: It is possible to train CycleGAN to learn the mapping between target landmarks and target identity. However, this method is not scalable nor suitable for our proposed problem. In order to translate images between a landmark domain and a face image domain, each CycleGAN needs to be trained separately for each identity. Besides, it can only generate results of the seen identity. 

{\bf StarGAN}: Compared to CycleGAN, StarGAN tackles the problem of solving multiple domains translation by fusing the target domain attributes with the given image by concatenating them channel-wise.  We established a StarGAN baseline that tries to straightforwardly adapt the StarGAN principles. We replaced the target attribute channels with a target landmarks image instead. Furthermore, we adjust its discriminator to output landmarks image instead of domain attributes similar to our video discriminator.
\subsubsection{Qualitative Results}
\begin{figure*}[t]
\begin{center}
  \includegraphics[width=\linewidth]{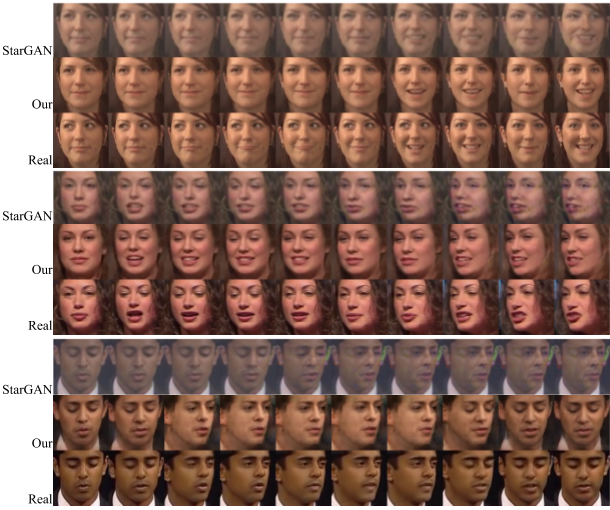}
\end{center}
   \caption{Qualitative comparison for video generation from landmarks between the results from StarGAN, our method, and the real videos.}
\label{fig:300vw}
\end{figure*}

Figure \ref{fig:300vw} shows the comparison between our method, a StarGAN baseline, and real videos. Each column is a video frame that is evenly sampled from the whole video sequence. Our method produces samples that are realistic as well as having closer colour to the real videos compared to StarGAN's results. Focusing on the last column on each of the examples, we can see that the StarGAN baseline tends to create artefacts and keep the background of the given image. We believe this is caused by the reconstruction loss proposed in the original implementation \cite{stargan}. Lastly, we can also observe that our model tends to produce results that have landmarks closer to real videos.
\subsubsection{Quantitative Results}
We quantitatively compare our result with a StarGAN baseline's by computing widely used reconstruction metrics: Peak Signal-to-Noise Ratios (PSNR) and Structural Similarity index (SSIM) \cite{psnrssim}. PSNR measures the difference between the produced results and the ground truth frames based on Mean Square Errors (MSE) and the maximum pixel value. SSIM is a method that quantifies the perceived quality between two images. For both of these metrics, larger values indicate better quality. We also evaluate the quality of the results based on their corresponding target landmarks. We use a state-of-the-art face alignment network \cite{jiankangfg2018} to estimate landmarks of the generated results and compare them with the ground truth landmarks. We then measure the error between landmarks with the point-to-point Root Mean Square (RMS) error normalized with the inter-ocular distance and reported them in the form of Cumulative Error Distribution (CED). From the calculated CED, we report Area Under Curve (AUC) value of the CED and also consider landmarks alignment Failure Rate (FR) for a maximum error at 0.1. Lastly, we report Face Matching Scores (FMS) based on Facesoft API \cite{FaceSoft}. The API calculate the probability of given two face images are from the same person. Hence, we report the FMS between ground truth video first frame and the produced videos. As can be seen in Table \ref{table:300vw}, our model achieves better result among all measurements indicating that our model produces more realistic results than the baseline, StarGAN. From AUC and FR, we can also see that our results indeed have landmarks perceived by \cite{jiankangfg2018} that are closer to the ground truth than the results from StarGAN. Last but not least, our method is the best at producing videos that preserve the identity achieving FMS of 76.28 \%.
\begin{table}
\begin{center}
\begin{tabular}{|c|c|c|c|c|c|}
\hline
Method & PSNR & SSIM & AUC & FR & FMS \\
\hline
\hline
\rule{0pt}{11pt} Real videos & - & - & 48.00 & 5.21 & 99.96\\
\hline
\rule{0pt}{11pt} StarGAN & 16.57 & 0.634 & 29.46 &  13.89 & 67.92\\
\hline
\rule{0pt}{11pt} Just $L^G_{img}$ & 17.08 & 0.693 & 28.88 & 15.02 & 36.15\\
\hline
\rule{0pt}{11pt} without $V$ & 17.75 & 0.705 & 31.12 &  10.32 & 51.35\\
\hline
\rule{0pt}{11pt} without $D_f$ & 18.00 & 0.714 & 31.47 &  9.62 & 66.36\\
\hline
\rule{0pt}{11pt} without $D_v$ & 18.08 & 0.715 & 30.02 &  12.72 & 67.70\\
\hline
\rule{0pt}{11pt} {\bf Our} & {\bf 18.12} & {\bf 0.716} & {\bf 33.97} & {\bf 7.79} & {\bf 76.28}\\
\hline
\end{tabular}
\end{center}
\caption{Quantitative results on 300VW dataset. The quality of videos are reported by PSNR and SSIM. Landmark accuracies are shown in AUC and FR at maximum error at 0.1. Face identities are tested with FSM.}
\label{table:300vw}
\end{table}
\begin{figure}
\begin{center}
  \includegraphics[width=\linewidth]{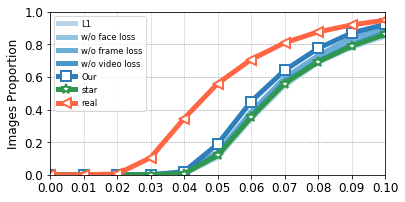}
\end{center}
   \caption{Our comparison of landmark localization results on the 300VW dataset. The results are reported as Cumulative Error Distribution of RMS point-to-point error normalized with interocular distance.}
\label{fig:ced}
\end{figure}
\subsubsection{Ablation Studies}
\begin{figure}[t]
\centering
\includegraphics[width=\linewidth]{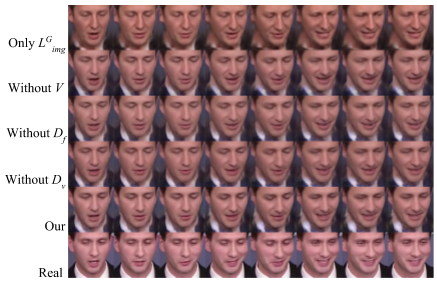}
\caption{Qualitative ablation studies on 300VW dataset with different combinations of losses}
\label{fig:ablation}
\end{figure}

Our framework consists of multiple networks and corresponding losses. In order to understand the values of each component, we repeat the training with different combination of losses: $L^G_{img}$, our without $V$, our without $D_f$, our without $D_f$, and our final loss. As shown in Table \ref{table:300vw}, we find that removing some or all of the components results in worse performance in all of the metrics. In particular, removing $V$ significantly reduce the ability of our model to preserve the identity. Similarly removing $D_v$ also negatively affects generated videos landmarks. Figure \ref{fig:ablation} demonstrates a qualitative comparison between losses.

\subsection{Video Synthesis based on Audio Inputs}
Here we demonstrate the applicability of our method by performing face video generation based on landmarks corresponding to audio inputs.

\subsubsection{Dataset}
We use GRID dataset \cite{grid} which has 33 speakers each pronouncing 1000 short phrases each containing 6 words from a limited dictionary. The dataset is divided into training and test sets according to \cite{dino} with a 70\% and 30\% proportion respectively. Since the dataset only contains the audio data associated with each video, we apply a face alignment method in \cite{jiankangfg2018} to establish the pseudo ground truths landmarks and resize each video frame to a common size $128 \times 128$.
Similarly, for training, we double the size of the training set by mirroring the training videos vertically.

\subsubsection{Baselines}

\hspace{0.35cm} {\bf Speech2Vid}: is a static method \cite{dinobaseline} that produces video frames from audio inputs using a sliding window, this is similar to our approach. This is a GAN-based method that utilizes a combination of an
$\ell_1$ loss and an adversarial loss on individual frames. 

{\bf SdfaGAN}: is a speech-driven facial animation framework that is based on temporal GANs. On top of deploying adversarial losses on individual frames, this method also exploits motion consistency that occurs in the produced video with a temporal adversarial loss.

\subsubsection{Qualitative Results}
\begin{figure*}
\centering
\includegraphics[width=\linewidth]{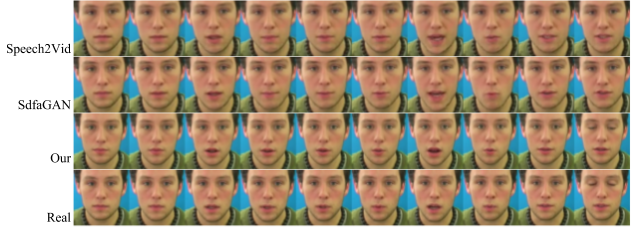}
\caption{Results of videos generated from audio inputs: Speech2Vid, SdfaGAN, our method, and ground truth videos.}
\label{fig:grid}
\end{figure*}

We show qualitative results in Figure \ref{fig:grid} in which we compare the results on the unseen identity. Each column represents a video frame evenly taken from the whole video. In a constrained setting, our generated results are visually indistinguishable from the real videos. In term of mouth and eye movements, we can see that our model generates videos that are realistic and mimics the real videos. Moreover, compared to two other baselines, our results are also consistently less blurry.

\begin{figure*}
\centering
\includegraphics[width=\linewidth]{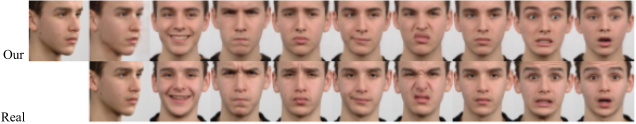}
\caption{Qualitative results on changing facial image emotions. Our method can alter emotion regardless of the head pose of source or target images.}
\label{fig:rafd}
\end{figure*}

\subsubsection{Quantitative results}
Apart from PSNR, SSIM, and FMS metrics, for this task, we also measure the accuracy of the spoken message using the Word Error Rate (WER) estimated by a pre-trained lip-reading network \cite{lipnet}. As shown in Table \ref{table:grid}, our model outperforms both baselines in term of the quality of the produced videos as measured by PSNR and SSIM. Considering WER metric, our method performs worse than SdfaGAN. However, this is expected since the method is specifically tailored to solve the problem. Interestingly, our method with $36.99\%$ WER surpasses the method proposed in \cite{dinobaseline} which achieves $37.2\%$ WER. Lastly, based on FMS, our method produces videos that preserve the most identities. This indicates that our method is sufficiently powerful and flexible to solve different face synthesis problems based on target landmarks.

\begin{table}
\begin{center}
\begin{tabular}{|c|c|c|c|c|}
\hline
Method & PSNR & SSIM & WER & FMS\\
\hline
\hline
\rule{0pt}{11pt} Real videos & - & - & 21.4\% & 99.99\\
\hline
\rule{0pt}{11pt} Speech2Vid & 27.39 & 0.831 & 37.2\% & 93.88\\
\hline
\rule{0pt}{11pt} SdfaGAN & 27.98 & 0.844 & {\bf 25.45}\% & 90.68\\
\hline
\rule{0pt}{11pt} {\bf Our} & {\bf 29.27} & {\bf 0.863} & 36.99\% & {\bf 97.08}\\
\hline
\end{tabular}
\end{center}
\caption{Quantitative results on GRID dataset. The quality of videos are reported by PSNR and SSIM measurements while the speech accuracy repoted by WER. Face identities are also reported with FSM.}
\label{table:grid}
\end{table}

\subsection{Changing Images Expressions}
We showcase another application of our methodology by changing an image's emotion based on landmarks. Unlike the method proposed in \cite{stargan}, our model does not assume the head pose of the source image nor the head pose of the target image (Figure \ref{fig:rafd} shows different head poses between source and target images).

\subsubsection{Dataset}
We use the Radboud Faces Database (RaFD) \cite{rafd} which contains 8,040 images from 73 participants each performing eight facial expressions simultaneously captured from five different angles. Because the dataset is not annotated with facial landmarks, we use \cite{jiankangfg2018} to extract the facial landmarks. The dataset also contains profile data in which we remove from our training set since our setting assume 68 points landmarks. We crop each image based on the pseudo ground truth landmark and resize them to a common scale at size $128 \times 128$.

\subsubsection{Qualitative Results}
We demonstrate qualitative results in Figure \ref{fig:rafd}. The first row shows our results and the second row displays corresponding real images. The first column contains a source image followed by a reconstructed image in the second columns. The following 8 columns are translated frontal facial images for 8 different emotions: happy, angry, sad, contemptuous, disgusted, neutral, fearful, and surprised respectively. Our results for changing facial images emotions look realistic and have landmarks closed to target landmarks. Denote that our model can arbitrarily change facial image emotion regardless of source or target facial landmarks. 


\section{Conclusions}
We proposed a very flexible methodology for editing facial images according to a target motion defined by a set of facial landmarks. Our methodology can be used for both facial expression/motion transfer, as well as the generation of an image sequence given a single facial image and the sequence of landmarks. We propose a novel way for training such a model so as to be robust to error accumulation. We demonstrate highly realistic video sequence creation driven by various poses and expressions. 

{\small
\bibliographystyle{ieee}
\bibliography{egbib}
}

\end{document}